# Tag Recommendation for Online Q&A Communities based on BERT Pre-Training Technique

1st Navid Khezrian
*Department of Computer Engineering*
*Sharif University of Technology*
Tehran, Iran
khezrian@ce.sharif.edu

2nd Jafar Habibi
*Department of Computer Engineering*
*Sharif University of Technology*
Tehran, Iran
jhabibi@sharif.edu

3rd Issa Anamoradnejad
*Department of Computer Engineering*
*Sharif University of Technology*
Tehran, Iran
i.moradnejad@gmail.com

*Abstract*— Online Q&A and open source communities use tags and keywords to index, categorize, and search for specific content. The most obvious advantage of tag recommendation is the correct classification of information. In this study, we used the BERT pre-training technique in tag recommendation task for online Q&A and open-source communities for the first time. Our evaluation on freecode datasets show that the proposed method, called TagBERT, is more accurate compared to deep learning and other baseline methods. Moreover, our model achieved a high stability by solving the problem of previous researches, where increasing the number of tag recommendations significantly reduced model performance.

## I. Introduction

Tags have recently emerged as a convenient way to organize and summarize user-generated content due to their ease of indexing and ease of user engagement. A tag is a keyword used to describe the subject of content and facilitates keyword classification and information search. The unique capability of tags to group and share information has changed the way people use information [1,2].

Tag recommendation is used to describe, summarize, and organize the content of objects. It not only improves the user experience by creating a space for specific, unambiguous Q & A, but can also increase the quality of the tags generated and indirectly improve the quality of data retrieval services so that the user gets better results after a query. In addition, recommendation effectiveness is highly important, considering that poor recommendations not only cause user dissatisfaction but also ultimately highly damage the performance of data recovery services [3].

Objects in open source communities, such as free-code or online social networking Q & A, include the name of the project or the title of the question, the description of the project, or the answer to the question and tags. When a developer posts a query on an online social network Q & A or shares a project in an open-source community, the site asks the user to select multiple tags for their content. Free-code allowed users to create more than 10 tags for each post [4].

In this study, we developed a BERT-based model for tag recommendation. BERT is designed to pre-train deep two-way shows of untagged texts. BERT is conceptually simple and empirically powerful [5]. We have implemented this model on free-code datasets and are running it on more datasets as well. Results of the experiments with TagCNN methods based on convolutional neural networks [6], TagRNN based on recurrent neural networks [7], TagHAN based on hierarchical attention networks [8], and TagRCNN based on recurrent convolution neural networks [9], which are widely used to classify texts, will be compared to traditional approaches to tagging: TagMulRec [10], EnTagRec [11], and FastTagRec [12]. Previous research has shown that not all deep learning methods can provide better results than traditional methods. Thus, our initial hypothesis focused on whether our proposed model, based on deep learning and using the BERT pre-learning technique, can provide better results than previous methods, especially the traditional methods of recommending tags on online social networking Q & A and open source communities.

The results of our experiments show that our model not only performed better than the traditional methods of tag recommendation but also achieved better results compared to the deep learning methods. It also solved the problem of previous researches, where increasing the number of tag recommendations significantly reduced their precision accuracy.

The rest of the article is organized as follows: The second part is the main body of the article, which reviews former studies in the field of tag recommendation and presents the proposed model. The third section describes the results of the evaluation of the proposed model, and finally, the fourth section presents the research conclusion.

## II. A Review of the Literature

Previous research on tag recommendation has been very extensive. Each study has tried to achieve a new model to improve the results by using new algorithms and methods as well as combining past solutions.

Zhou et al. [4] in 2019 compared traditional methods of tag recommendation with deep learning algorithms on different datasets. They compared the EnTagRec, TagMulRec, and FastTagRec methods, which are considered as the best traditional methods in tagging, with the TagCNN, TagRNN, TagHAN, and TagRCNN algorithms. Their results showed that TagRCNN, TagCNN was superior to traditional methods in all their datasets, but other deep learning methods could not achieve the desired results.

By examining the methods used in tag recommendation, one can conclude that all the approaches introduced can be classified into six main categories that Belam et al. [3] discussed in 2016.

### A. Co-Occurrence-Based Methods

Co-occurrence-based methods use associative rules and tags assigned to the previous object sets to recommend tagging for a new object. Hence, this method requires a large set of tagged data.

### B. Graph-Based Methods

Graph-based methods extract recommended tags from the object's neighborhood. In these approaches, nodes represent objects, and edges show the similarity relationship between objects.

### C. Matrix Factorization-Based Methods

Matrix factorization-based methods apply the tag as a model matrix and apply dimensional methods to that matrix. They aim to recommend tags by predicting relationships between users, tags, and objects. This approach works well when there are datasets about users, tags, and objects.

### D. L2R-Based Methods

L2R-based methods recommend tags based on automatic learning of the training set as a feature vector. The purpose is to automatically combine the characteristics and qualitative variables of the tag and produce a model that ranks these characteristics in a score or position according to the purpose of the recommendation.

### E. Content-Based Methods

Content-based methods derive tag recommendation based on the target content and related features or characteristics of the target user.

### F. Clustering-Based Methods

Clustering-based methods are used to group objects and tags, and representative tags from the target object cluster are recommended. Clustering is an interesting strategy to diminish the dimensions of the problem. Such methods use the relationships between clusters to create representative tags.

## III. PROPOSED SOLUTION

In this section, we will first state the tag recommendation problem for textual datasets, then the TagBERT method for solving this problem.

In the case of tag recommendation, the main question is how to automatically recommend a set of tags for a new object, in the case of a set of tagged objects. To answer this question, we created the Tag-BERT model.

### A. TagBERT

Figure 1 shows the TagBERT model for tag recommendation. The main steps of our proposed model are as follows:

1. *We used the data pre-processing rules in [4,13,14,15] to get a set of tags.*

2. *Next, for each object, we have an array of N tags that specify what tags that object has, in a binary approach.*

3. *In this step, we will perform pre-training and sentence embedding operations using the BERT module, which required a series of preprocesses. First, we divide our objects into several sections and equalize the length of the objects in each section. Then, the Tokenaization operation is performed on each section. At this stage, the BERT module assigns a number to each of our tokens based on its training process.*

4. *To increase the accuracy of our model, we give the output of our BERT module to a text model. In this model, each of our objects passes through a CNN, whose region size is set to (4, 3, 2) and has 50 filters for each region size, as well as a DNN layer with 256 units. After performing the necessary processes in this text model, a number between 0 and 1 is assigned to the components of each tag from the set of tags to the desired object.*

5. *In this step, we chose a limit for assigning tags to objects by trial and error and repeating the steps. The value of this variable in our model is 0.92. At this stage, tags that are more likely to be this way will be recommended to the object; otherwise, no recommendation will be made.*

## IV. RESULTS

In order to be consistent with previous studies [12,14,15,16] in our experiments, we performed the model 10 times and reported the average of the results as the final result.

We randomly selected 10,000 objects from the data set to evaluate the model as a test set (V) and used the remaining objects for the training process.

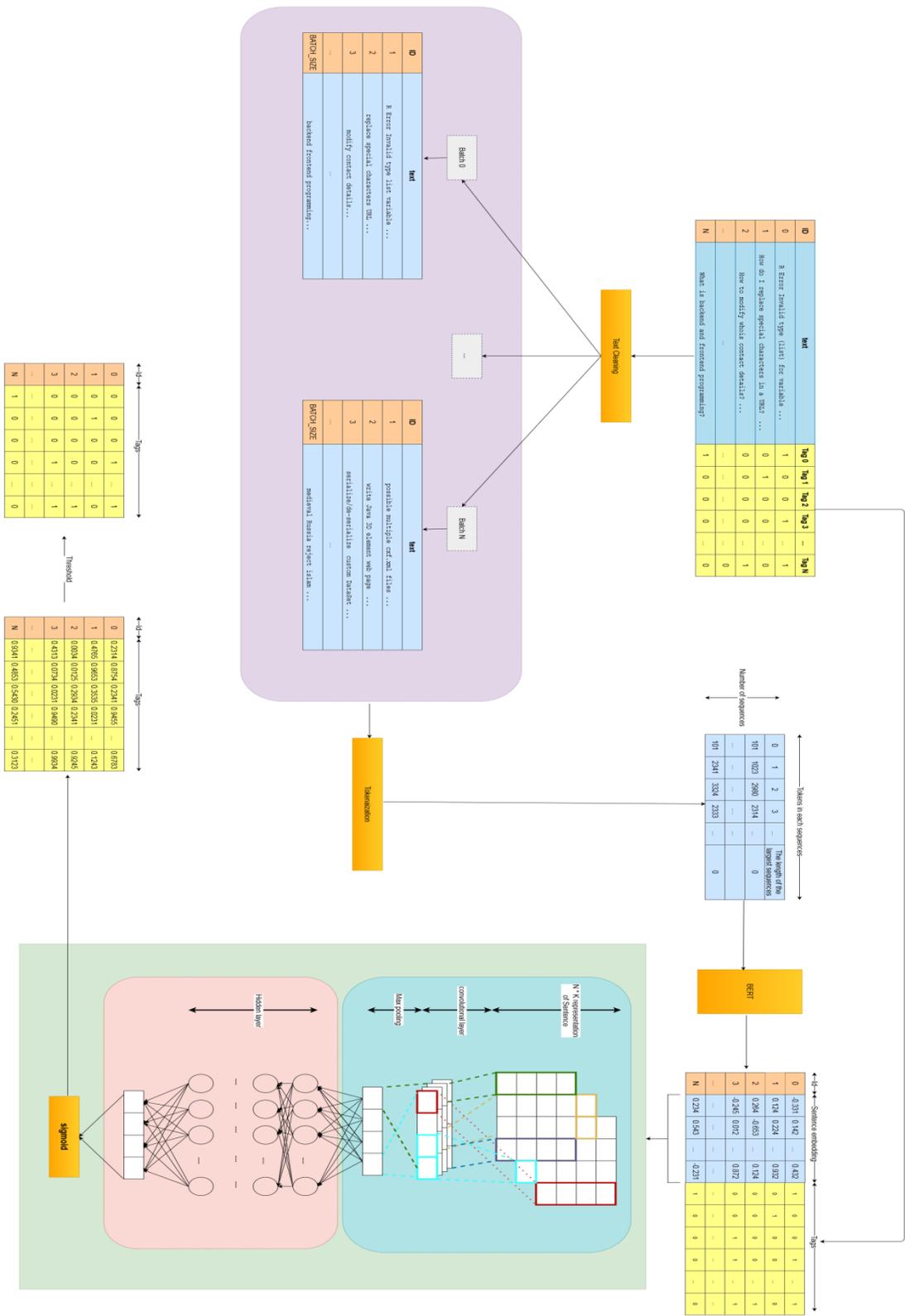

Figure 1: The main stages of the TagBERT model

For each object that is part of our test set ($O_i \in V$), a maximum of K tags is recommended to create the recommended tag set ($TR_i^K$) if our model decides that the item qualifies to have K tags.

All tests were performed on a 64-bit Intel Core i7 3.6G computer with 64 GB of RAM with Ubuntu 16.04. We used the Tensorflow3 open-source software library to implement the model.

To match the experiments reported in [12,14,15], we also used Recall@K, Precision@K, and F1-score@K to evaluate the results of our experiments.

To calculate Recall@Ki from the set of recommended tags and the set of real tags of an object ($O_iT$), we use equation (1). we use equation (2) to calculate the Recall@K of all the objects in the test set (V).

$$\text{Recall@K}_i = \begin{cases} \text{Recall@K}_i = \frac{|TR_i^K \cap O_iT|}{K}, & |O_iT| > K \\ \text{Recall@K}_i = \frac{|TR_i^K \cap O_iT|}{O_iT}, & |O_iT| \leq K \end{cases} \quad (1)$$

$$\text{Recall@K} = \frac{\sum_{i=1}^{|V|} \text{Recall@K}_i}{|V|} \quad (2)$$

Equation (3) is used to calculate Precision@Ki from the set of recommended tags and the set of real tags of an object ($O_iT$). Also, Equation (4) is used to calculate Precision@K for all objects in the test set (V).

$$\text{Precision@K}_i = \frac{|TR_i^K \cap O_iT|}{K} \quad (3)$$

$$\text{Precision@K} = \frac{\sum_{i=1}^{|V|} \text{Precision@K}_i}{|V|} \quad (4)$$

Finally, to calculate the F1-score@Ki criterion, we use the combination of Recall@Ki and Precision@Ki, according to formula (5), and using formula (6), we calculate F1-score@K for all objects in the test set (V).

$$\text{F1-score @K}_i = 2 \cdot \frac{\text{Precision@K}_i \cdot \text{Recall@K}_i}{\text{Precision@K}_i + \text{Recall@K}_i} \quad (5)$$

$$\text{F1-score @K} = \frac{\sum_{i=1}^{|V|} \text{F1-score @K}_i}{|V|} \quad (6)$$

We apply the tag recommendation model to the preprocessed data for training and use the test set (V) to evaluate the model based on the criteria introduced (K=10). The results obtained in our experiments are summarized in Table 1 for the free-code data sets. As you can see in Table 1, our proposed model in the F1-score@10 criterion has performed about 10% better than the best traditional method.

F1-score@10 is a vital criterion in the evaluation because it is calculated from the combination of Precision and Recall. Compared to the best method of deep learning models, our method has made a small improvement as well. It should be noted that in tag recommendation problems, because of the very high number of classes and detailed categorization of objects, increasing accuracy for achieving high levels in the Precision and F1-score criteria is truly challenging

Table 1: Comparison of TagBERT performance with traditional methods and deep learning based on F1-score@10, Precision@10, and Recall@10 criteria for free-code datasets

| models | F1-score@10 | Precision@10 | Recall@10 |
|---|---|---|---|
| TagMulRec | 36.4 | 24.5 | 75.8 |
| EnTagRec | 36 | 23.9 | 77.3 |
| FastTagRec | 33.2 | 21.9 | 82 |
| TagCNN | 45.3 | 29.7 | **94.9** |
| TagRNN | 20.8 | 13.8 | 41.6 |
| TagHAN | 23.3 | 15.1 | 48.4 |
| TagRCNN | 39.2 | 25.8 | 81.6 |
| **TagBERT** | **46.5** | **40.25** | 64.42 |

Our proposed model is less accurate in the Recall criterion than other models. The reason for this is that we have paid attention to all the criteria, and in our proposed model, we have tried to provide good accuracy in the Precision criterion as well. Consequently, the parameters in our proposed model have been adjusted to provide acceptable results in both Recall and Precision criteria and to solve the problems of previous studies, where Precision criteria decreased significantly as the number of recommendations increased.

As you can see in Table 2, in our proposed model (TagBERT) with increasing the number of tag recommendations from 5 tags to 10, the Precision criterion has not decreased much, but in other models, it has dropped significantly.

In this section, we are still evaluating the results on other datasets.

Table 2: Comparison of TagBERT performance with traditional methods and deep learning by increasing the number of tag recommendations (5 to 10) based on Precision criteria for free-code datasets

| Name | Precision@5 | Precision@10 |
|---|---|---|
| TagMulRec | 38.3 | 24.5 |
| EnTagRec | 37.9 | 23.9 |
| FastTagRec | 34.3 | 21.9 |
| TagCNN | 53 | 29.7 |
| TagRNN | 21.1 | 13.8 |
| TagHAN | 21.8 | 15.1 |
| TagRCNN | 47.8 | 25.8 |
| **TagBERT** | 41.83 | **40.25** |

## V. CONCLUSION

In this study, our goal was to present a new deep learning model based on the BERT pre-training technique, which was used for the first time to assist tag recommendation in online social networking Q & A. Not only did TagBERT perform better than other deep learning and traditional tagging methods, but it also solved the problem of previous studies, where Precision was significantly reduced by increasing the number of tag recommendations. In our model, the increase of recommendations will not change Precision much.

We believe that deep learning can be of great help in improving and resolving problems in online social networking Q & A. Therefore, as our study complemented previous research, various deep learning strategies should also be carefully evaluated and analyzed in order to get the most optimal response for tag recommendation.